\documentclass[sigconf]{acmart}
\usepackage{balance}

\usepackage[ruled,vlined,linesnumbered]{algorithm2e}
\usepackage{bm}
\usepackage{amsfonts}
\usepackage{multirow}
\usepackage{subcaption}
\usepackage{diagbox}
\usepackage{subscript}
\usepackage{siunitx}

\def\etal{\emph{et al. }}
\newcommand{\figref}[1]{Figure \ref{#1}}
\newcommand{\tabref}[1]{Table \ref{#1}}
\newcommand{\eqnref}[1]{Eq.\ref{#1}}

\newcommand{\com}{\text{,}}

\AtBeginDocument{%
  \providecommand\BibTeX{{%
    \normalfont B\kern-0.5em{\scshape i\kern-0.25em b}\kern-0.8em\TeX}}}

\copyrightyear{2020} 
\acmYear{2020} 
\setcopyright{acmcopyright}
\acmConference[MM '20]{Proceedings of the 28th ACM International Conference on Multimedia}{October 12--16, 2020}{Seattle, WA, USA}
\acmBooktitle{Proceedings of the 28th ACM International Conference on Multimedia
(MM '20), October 12--16, 2020, Seattle, WA, USA}
\acmPrice{15.00}
\acmDOI{10.1145/3394171.3414064}
\acmISBN{978-1-4503-7988-5/20/10}

\acmSubmissionID{1678}

\begin{document}
\fancyhead{}
\title{Query Twice: Dual Mixture Attention Meta Learning \\ for Video Summarization}

\author{Junyan Wang$^{1}$, Yang Bai$^{2}$, Yang Long$^{3}$,} 
\author{Bingzhang Hu$^{2}$, Zhenhua Chai$^{1}$, Yu Guan$^{2}$, Xiaolin Wei$^{1}$}
\affiliation{%
  \institution{$^{1}$Vision Intelligence Center, Meituan-Dianping Group, Beijing, China}
  \institution{$^{2}$OpenLab, Newcastle University, Newcastle upon Tyne, UK}
  \institution{$^{3}$Department of Computer Science, Durham University, Durham, UK}
  \institution{\{wangjunyan04, chaizhenhua, weixiaolin02\}@meituan.com,}
  \institution{\{y.bai13, bingzhang.hu, yu.guan\}@newcastle.ac.uk, yang.long@ieee.org}
}








\renewcommand{\shortauthors}{Wang et al.}


\begin{abstract}
Video summarization aims to select representative frames to retain high-level information, which is usually solved by predicting the segment-wise importance score via a softmax function. However, softmax function suffers in retaining high-rank representations for complex visual or sequential information, which is known as the \textit{Softmax Bottleneck} problem. In this paper, we propose a novel framework named Dual Mixture Attention (DMASum) model with Meta Learning for video summarization that tackles the softmax bottleneck problem, where the Mixture of Attention layer (MoA) effectively increases the model capacity by employing twice self-query attention that can capture the second-order changes in addition to the initial query-key attention, and a novel Single Frame Meta Learning rule is then introduced to achieve more generalization to small datasets with limited training sources. Furthermore, the DMASum significantly exploits both visual and sequential attention that connects local key-frame and global attention in an accumulative way. We adopt the new evaluation protocol on two public datasets, SumMe, and TVSum. Both qualitative and quantitative experiments manifest significant improvements over the state-of-the-art methods.
\end{abstract}

\ccsdesc[500]{Computing methodologies~Video summarization}

\keywords{video summarization, attention network, meta learning}

\maketitle

{\fontsize{8pt}{8pt} \selectfont
\textbf{ACM Reference Format:}\\
Junyan Wang, Yang Bai, Yang Long, Bingzhang Hu, Zhenhua Chai, Yu Guan and Xiaolin Wei. 2020. Query Twice: Dual Mixture Attention Meta Learning for Video Summarization. In \textit{Proceedings of the 28th ACM International Conference on Multimedia (MM’20), October 12--16, 2020, Seattle, WA, USA.} ACM, New York, NY, USA, 9 pages. https://doi.org/10.1145/3394171.3414064 }

\section{Introduction}
\begin{figure}[t]
 \centering
 \includegraphics[width=\columnwidth]{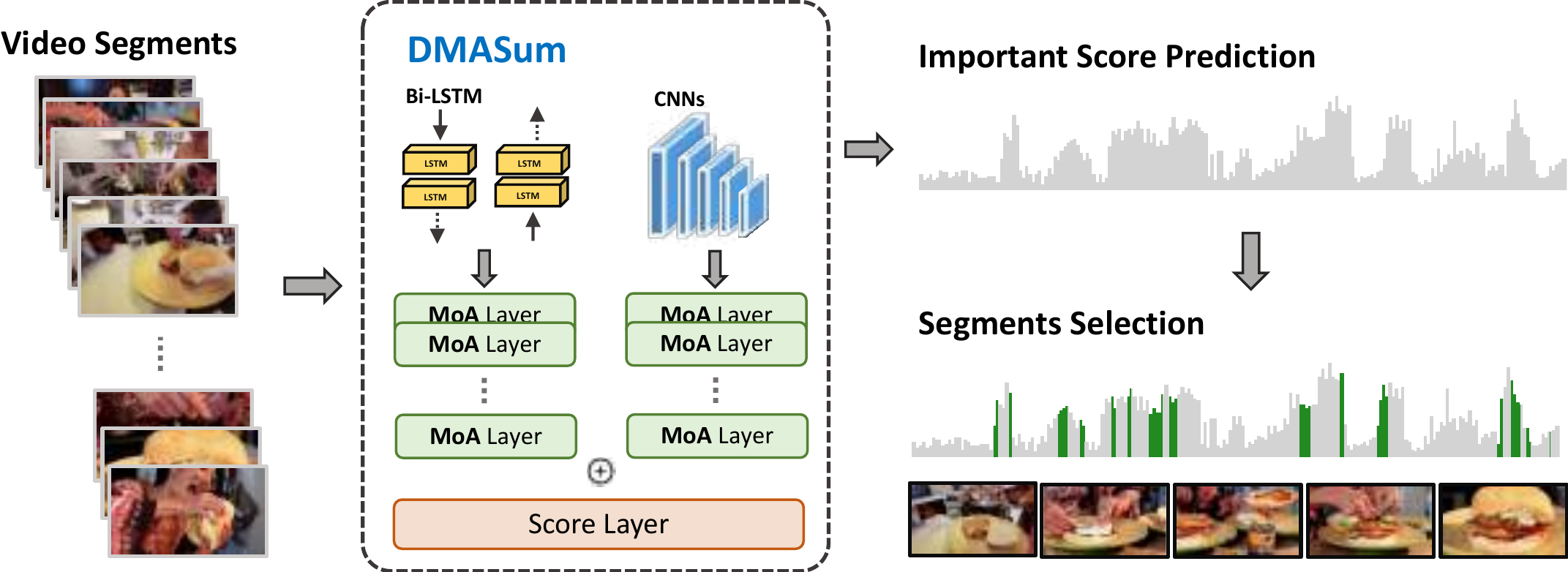}  
 \caption{An illustration of the video summarization task using our proposed DMASum. Each gray bar represents the predicted important score of a segment and green bars denote the key-segments in the summarized video. Highlights of DMASum include Visual-sequential Dual Channels, Stacked MoA modules.}
 \label{fig:overview}
 \end{figure}

With the tremendous growth of video materials uploaded to various online video platforms like YouTube, automatic video summarization has received increasing attention in recent years.
The summarized video can be used in many scenarios such as fast indexing and human-computer interaction in a light and convenient fashion. The main objective of video summarization is to shorten a whole video into summarized frames while preserving crucial plots. 
One of the mainstream directions focuses on key-frames summarization \cite{gygli2014summe} is illustrated in Fig. \ref{fig:overview} 
A video is first divided into 15-second segments, and the problem is modeled as an importance score prediction task to select the most informative segments.
 
The nature of video summarization task encourages a line of research \cite{wei2018sasum,mahasseni2017sumgan,yuan2019cyclesum,jung2019csnet} focusing on unsupervised learning methods. 
Besides, \cite{zhou2018dlsum} applied deep reinforcement learning with a diversity-representativeness reward function for the generated summary; 
Currently, the most popular benchmarks are SumMe \cite{gygli2014summe} and TVSum \cite{song2015tvsum}. Otani \etal \cite{otani2019rethinking} proposed to evaluate the methods by using the rank-order correlation between predicted and human-annotated importance scores. These key evaluation matrices measure agreements between generated summaries and reference summaries. Therefore, supervised methods \cite{gong2014seqdpp,zhang2016vslstm,wei2018sasum,zhao2018hsarnn} are still very important for investigating essential technical questions because they can directly compare against human-annotated scores as ground truth. 
One of the mainstream directions focuses on key-frames summarization \cite{gygli2014summe} is illustrated in Fig. \ref{fig:overview}.

The challenges for supervised key-frames summarization are two-fold.
First, the importance scores are very subjective and highly related to human perception. Second, the annotations are expensive to be obtained; thus, the model should be able to cope with limited labeled data while retaining high generalization. 
These are not only unsolved questions for video summarization but also essential for many other research domains. 
To this end, this paper proposes a new framework, namely the Dual Mixture Attention model (DMASum) that aims to achieve \textbf{1) }human-like attention by adopting cutting-edge self-attention architecture and takes both visual and sequential information into a unified process; and \textbf{2)} high-level semantic understanding of the whole content by incorporating a novel meta learning module to maximally exploit the training data and improve the model generalization. 

The proposed framework manifested promising results in our early experiments. However, the early implementation reflected two major technical challenges. The first is known as the \textit{Softmax Bottleneck} problem associated with the self-attention architecture. Both theoretical and empirical evidences in this paper show that traditional softmax function does not have the sufficient capability to retain high-rank representation for long and complex videos. 
To this end, we propose a \textit{Query Twice} module by adding self-query attention to query-key attention. The Mixture of Attention layer can then compare the two attentions to capture the second-order changes and increase the model capability.
The second problem is that the most common meta learning strategy does not naturally fit the video summarization task. We propose a Single-video Meta Learning rule to refrain the learner tasks so as to purify the meta learner updating processes. To summarize our contributions:
\begin{itemize}
    \item To our best knowledge, this is the first paper that successfully introduces self-attention architecture and meta learning to jointly process dual representations of visual and sequential information for video summarization.
    \item We provide in-depth theoretical and empirical analyses of the Softmax Bottleneck problem when applying attention model to video summarization task. And a novel self-query module with Mixture-of-Attention is provided as the solution to overcome the problem effectively.
    \item We explore the meta learning strategy, and a Single-Video Meta Learning rule is particularly designed for video summarization tasks.
    \item Quantitatively and qualitatively experiments on two datasets: SumMe \cite{gygli2014summe} and TVSum \cite{song2015tvsum} demonstrate our superior performance over the state-of-the-art methods. More impressively, our model achieves human annotator level performance under new protocols of Kendall's $\tau$ correlation coefficients and Spearman's $\rho$ correlation coefficients. The groundbreaking results suggested that our DMASum has effectively modeled human-like attention.
\end{itemize}

\section{Related Work}

\noindent\textbf{Video summarization. }
Video, as a media containing complex spatio-temporal relationship of visual contents, has a wide range of applications \cite{wu2019long,feichtenhofer2019slowfast,zhu2018towards,wang2019fast,wang2019order,dwibedi2019temporal}.
However, because of its huge volume, video summarization is to compress such huge volume data into its light version while preserving its information.
Early works have presented various solutions to this problem, including storyboards \cite{lu2013story,gong2014seqdpp,gygli2015submodular,lu2013story} and objects \cite{liu2010objectsummary,lee2012discovering,zhenhua}. 
LSTM-based deep learning approaches are proposed for both supervised and unsupervised video summarization in recent years. Zhang \etal \cite{zhang2016vslstm} proposed a bidirectional LSTM model to predict the importance score of each frame directly, and this model is also extended with determinantal point process \cite{kulesza2012dpp}. Mahasseni \etal \cite{mahasseni2017sumgan} specified a generative adversarial framework that consists of the summarizer and discriminator for unsupervised video summarization. The summarizer is an auto-encoder LSTM network for reconstructing the input video, and the discriminator is another LSTM network for distinguishing between the original video and its reconstruction.
Meanwhile, based on the observation of Otani \etal \cite{otani2019rethinking}, they propose another evaluation approach as well as a visualization of correlation between the estimated scoring and human annotations. 

\begin{figure*}[t]
\centering
\includegraphics[width=0.85\textwidth]{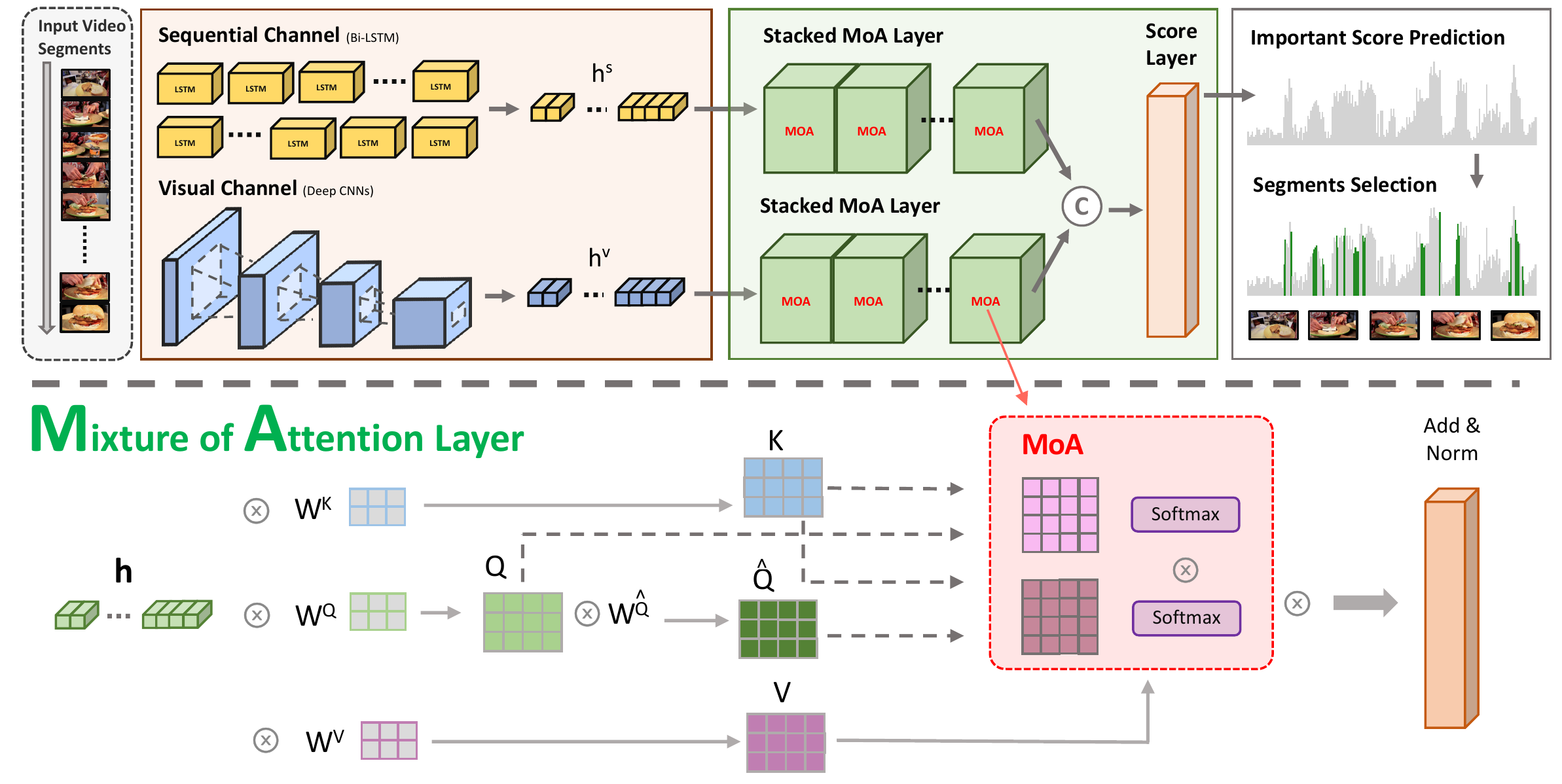} 
\vspace{-2ex}
\caption{The overall architecture of our DMASum is shown as the top figure, which consists of a sequential channel and a visual channel and stacked MoA layers. The bottom part shows the structure of the Mixture of Attention layer.}
\label{fig:architecture}
\end{figure*}

\noindent\textbf{Attention-based Models. } 
The attention mechanism was born to help memorize long source sentences in neural machine translation \cite{bahdanau2014attention1}. 
Rather than building a single context vector out of the translation encoder, the attention method is to create shortcuts between the context vector and the entire input sentence, then customize the weights of these shortcut connections for each element. 
The Transformer \cite{vaswani2017transformer}, without a doubt, is one of the most impressive works in the machine translation task. 
The model is mainly built on self-attention layers, also known as intra-attention, and the self-attention network is relating different positions of the same input sequence. 
Many recent works have applied self-attention to a wide range of video-related applications, such as video question answer \cite{li2019vqa} and video captioning \cite{zhou2018caption}.
Particularly for the video summarization task, Ji \etal \cite{ji2019autoae} proposed an attention-based encoder-decoder network for selecting the key shots. 
He \etal \cite{he2019unsupervised} proposed an unsupervised video summarization method with attentive conditional Generative Adversarial Networks.

\noindent\textbf{Meta Learning. } 
Meta learning, also known as learning to learn, aims to design a model that can be learned rapidly with fewer training examples.
Meta learning usually used in few-shot learning \cite{finn2017maml,nichol2018reptile} and transfer learning \cite{wang2016ltol}.
Finn \etal \cite{finn2017maml} propose a Model Agnostic Meta Learning (MAML) which is compatible with any model trained with gradient descent and applicable to a variety of different learning problems, including classification, regression, and reinforcement learning. 
Like MAML, the work of Nichol \etal \cite{nichol2018reptile} proposed a strategy which repeatedly sampling and training a single task, then moving the initialization towards the trained weights on that task.
Recently, meta learning methods have been applied in a few video analysis tasks.
Especially in video summarization, Li \etal \cite{li2019MetaLTDVS} proposed a meta learning method that explores the video summarization mechanism among summarizing processes on different videos.

\section{The Proposed Approach}
Video summarization is modeled as a sequence labeling (or sequence to sequence mapping) problem. 
Given a sequence of video frames, the task is to assign each frame an importance score based on which key-frames can be selected. Existing sequence labelling approaches include deep sequential models such as LSTM \cite{zhang2016vslstm,zhang2018rssum}, attention model \cite{ji2019autoae}. However, the key difficulty is to learn the frame dependencies within the video and capturing the internal contextual information of the video. Considering video is a highly context-dependent source that shares many similar properties in sentences. As the outstanding performance of the Transformer \cite{vaswani2017transformer}, we introduce the self-attention structure that has been widely used in natural language processing (NLP) as our architecture basis. Both visual and sequential representations are considered in order to model complex human-like attention and better match the subjective annotations. Also, the motivation of meta learning aims to improve the model generalization when training sources are insufficient due to expensive human annotations. An overview of the proposed video summarization architecture 
and the details of the Mixture of Attention layer that are illustrated in \figref{fig:architecture}.

\subsection{Architecture Design}
\noindent\textbf{Dual-representation Learning:} For the video summarization task, we introduce both visual and sequential channels as the input. The visual channel (deep CNNs) extracts visual features $\boldsymbol{H}^{v} = \{\boldsymbol{h}^{v}_t \}^T_{t=1}$ from each video frame image. Based on the extracted visual features, the sequential features $\boldsymbol{H}^{s} = \{\boldsymbol{h}^{s}_t \}^T_{t=1}$ is obtained by the sequential channel (bidirectional LSTM network) and consists of the dual-channel feature $\boldsymbol{H} \in \{\boldsymbol{H}^{v}, \boldsymbol{H}^{s}\}$. The dual representation is critical to model complex human-like attention and can link frame-wise attention to the overall story line.

\noindent\textbf{The Attention Module:} Taking a feature sequence $\boldsymbol{H} = \{\boldsymbol{h}_t \}^T_{t=1} \in \mathbb{R}^{D \times T}$ extracted from the video as input, the attention network can re-express each $\boldsymbol{h}^{*}_t$ within input $\boldsymbol{H}$ by utilizing weighted combination of the entire neighborhood from $\boldsymbol{h}_1$ to $\boldsymbol{h}_T$, where $D$ is the feature dimension and $T$ is number of frames within a video. In concreteness, the attention network first linearly transforms $\boldsymbol{H}$ into $\boldsymbol{Q} = \boldsymbol{W}^Q\boldsymbol{H}^{*}$, $\boldsymbol{K} = \boldsymbol{W}^K\boldsymbol{H}^{*}$ and $\boldsymbol{V} = \boldsymbol{W}^V\boldsymbol{H}^{*}$, where $\boldsymbol{Q} = \{\boldsymbol{Q}_t \}^T_{t=1} \in  \mathbb{R}^{D_{a} \times T}$, $\boldsymbol{K} = \{\boldsymbol{K}_t \}^T_{t=1} \in  \mathbb{R}^{D_{a} \times T}$ and $\boldsymbol{V} = \{\boldsymbol{V}_t \}^T_{t=1} \in  \mathbb{R}^{D_{a} \times T}$ are known  as Queries, Keys and Values vectors, respectively and $D_a$ represents the attention feature size, and $\boldsymbol{W}^Q, \boldsymbol{W}^K, \boldsymbol{W}^V \in \mathbb{R}^{D_{a} \times D}$ are the corresponding learnable parameters. 
$\boldsymbol{K}$ is employed to learn the distribution of attention matrix on condition of the query matrix $\textbf{Q}$, and $\boldsymbol{V}$ is used to exploit information representation.
Thus the scaled dot-product attention $\boldsymbol{A}$ is defined as:
\begin{equation}
      \mathcal{F}_{Scale}(\boldsymbol{K},\boldsymbol{Q}) = \frac{\boldsymbol{K}^T\boldsymbol{Q}}{\sqrt{D_a}}~\com
      \label{eq:dot-product}
\end{equation}
\begin{equation}
      \boldsymbol{A} = \mathcal{F}_{Softmax}(\boldsymbol{K},\boldsymbol{Q}) = \frac{\text{exp}(\mathcal{F}_{Scale}(\boldsymbol{K},\boldsymbol{Q}))}{\sum_{t=1}^{T}\text{exp}\mathcal{F}_{Scale}(\boldsymbol{K},\boldsymbol{Q})}~\com
      \label{eq:alpha}
\end{equation}
where $\boldsymbol{A} \in \mathbb{R}^{T \times T}$ and we consider $\boldsymbol{A}$ as the distribution of attention matrix on condition of the query matrix $\textbf{Q}$. In \eqnref{eq:dot-product}, due to the large degree of high dimensional $\boldsymbol{K}^T\boldsymbol{Q}$, scaling factor $\frac{1}{\sqrt{D_a}}$ is used to prevent the potential small gradient suffered by softmax. The output of attention network is:
\begin{equation}
   \boldsymbol{Z} = \boldsymbol{V}\boldsymbol{A}.
\end{equation}

After applying the attention module to both channels, We concatenate their outputs and feed into a score layer, which consists of multiple fully-connected layers ended with a sigmoid function.
The score layer predicts the importance score $\hat{s}$ is sampled as:
\begin{equation}
    \begin{aligned}
      \hat{\boldsymbol{S}} &= \mathcal{F}_{Score}(\mathcal{F}_{Concat}(\boldsymbol{Z}^v,\boldsymbol{Z}^s))\com \\
    \end{aligned}
\label{eq:score}
\end{equation}
where $\mathcal{F}_{Score}$ denotes the score layer and $\mathcal{F}_{Concat}$ in this paper means concatenation operation on different channels.

\noindent\textbf{Overall Objective Function. }
We intend to treat the outputs as the importance scores of the whole video frames in this work. 
Thus, we simply employ the mean square loss  $\mathcal{L}$ between the ground truth importance scores and the predicted importance scores.
\begin{equation}
       \mathcal{L} =  {\frac {1}{T}}\sum _{i=1}^{T}(s_{i}-{\hat {s_{i}}})^{2}\com
\end{equation}


\subsection{The Softmax Bottleneck}
Almost all existing attention models follow the original pipeline from NLP tasks using the softmax function Eq. \eqref{eq:alpha} to compute the attention. However, this section identifies the key limitation of softmax function for video summarization. It can be considered that the attention distribution is a finite set of pairs of a context and its conditional distribution $\mathcal{V} = \{(c_{1}, P^{*}(X \vert c_{1})) , \dots ,(c_{1}, P^{*}(X \vert c_{T}))\}$, where $\mathcal{X} = \{x_{1},x_{2},\dots,x_{N}\}$ denotes T compatible keys in the video $\mathcal{V}$ and $\mathcal{C} = \{c_{1},c_{2},\dots,c_{N}\}$ denotes the contexts.
It is assumed $P^{*} > 0$ and $\boldsymbol{A}^{*}$ represents the true attention distribution. Thus the true attention distribution in \eqref{eq:alpha} can be re-formulated as:
\begin{equation}
      \boldsymbol{A}^{*} = \left[
 		\begin{matrix}
   			log P^{*}(x_{1} \vert c_{1}) & log P^{*}(x_{2} \vert c_{1}) & \cdots & log P^{*}(x_{T} \vert c_{1})\\
   			log P^{*}(x_{1} \vert c_{2}) & log P^{*}(x_{2} \vert c_{2}) & \cdots & log P^{*}(x_{T} \vert c_{2})\\
  			\vdots & \vdots & \ddots & \vdots\\
  			log P^{*}(x_{1} \vert c_{T}) & log P^{*}(x_{2} \vert c_{T}) & \cdots & log P^{*}(x_{T} \vert c_{T}).\\
  			\end{matrix}
  		\right]  
      \label{eq:matrix}
      \medskip
\end{equation}

The objective of attention model is to learn the conditional attention distribution $P_{\theta}(\mathcal{X} \vert \mathcal{C})$ parameterized by $\theta$ to match the true attention distribution $P^{*}(\mathcal{X} \vert \mathcal{C})$. It can be seen that the attention distribution problem is now turned into a \textbf{matrix factorization problem}. Since $\boldsymbol{A}$ is a matrix with size $N \times N$, the rank of learned attention distribution $\boldsymbol{A}$ is upper bounded by the embedding size $d$. If $d < rank(\boldsymbol{A}^{*}) - 1$, for any model parameter $\theta$, there exists a context $c$ in $\mathcal{V}$ such that $P_{\theta}(\mathcal{X} \vert \mathcal{C}) \neq P^{*}(\mathcal{X} \vert \mathcal{C})$. This is so called \textbf{\textit{Softmax Bottleneck}} which reflects the circumstance when softmax function does not have the capacity to express the true attention distribution when $d$ is smaller than $rank(\boldsymbol{A}^{*}) - 1$.
In the contexts of video summarization, the log probability matrix $\boldsymbol{A}$ becomes a high-rank matrix when the visual contents are complex and the changes between frames are severe. For example, cooking may contain multiple repetitive actions than eating. While humans can intuitively assign equal importance to both of the actions, the former one actually results in a much higher rank in the representation matrix. The softmax function may compromise features from rich content to maintain consistency.

\begin{figure}[t]
\centering
    \begin{subfigure}[b]{0.24\textwidth}
        \centering
        \includegraphics[width=0.95\textwidth]{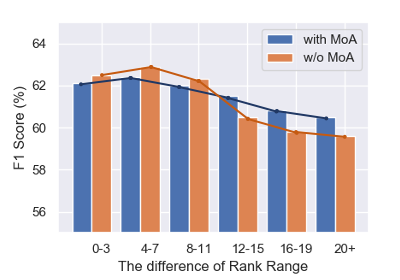}
        \caption{Visual Channel}
    \end{subfigure}%
    \begin{subfigure}[b]{0.24\textwidth}
        \centering
        \includegraphics[width=0.95\textwidth]{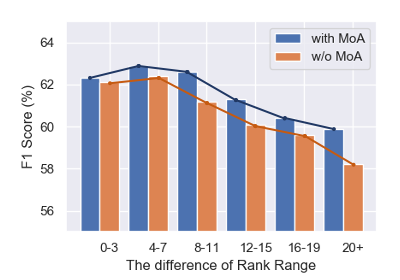}
        \caption{Sequential Channel}
    \end{subfigure}%
    \smallskip
    \newline
    \begin{subfigure}[b]{0.24\textwidth}
        \centering
        \includegraphics[width=0.95\textwidth]{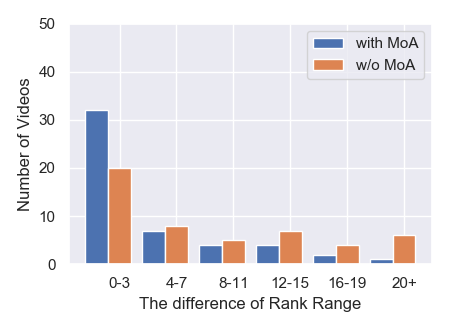}
        \caption{Visual Channel}
    \end{subfigure}%
    \begin{subfigure}[b]{0.24\textwidth}
        \centering
        \includegraphics[width=0.95\textwidth]{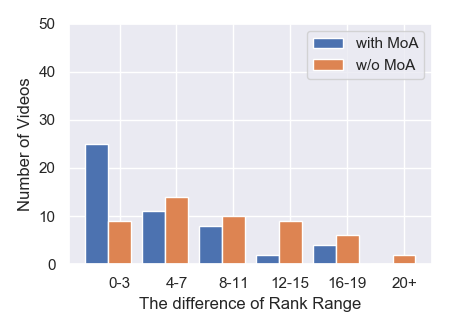}
        \caption{Sequential Channel}
    \end{subfigure}%
    \caption{Averaged F1-score (\%) and Number of videos with respect to the rank difference $\mathcal{D}$ in TVSum dataset.Blue and Orange bars compare our MoA against traditional softmax. }
\label{fig:rank observation}
\end{figure}

In \figref{fig:rank observation} we empirically verify such a Softmax Bottleneck problem can degrade the performance severely. We choose the TVSum dataset and calculate the difference $\mathcal{D} = T - rank(\boldsymbol{A})$, where $T$ denotes the video length. This is because video lengths are not consistent so we only consider the difference between the actual rank and the full rank $T$. Lower difference values indicate the attention layer, after softmax, can retain high rank with minimum redundancy. On the other hand, Higher difference values mean the attention matrix of the whole video is low-rank. It can be due to the input video is not complex, e.g. no movement and the background is monotonous. But for most of the cases, the low-rank attention matrix is often resulted by key information missing due to long videos with high complexities. The statistics are collected from attention matrices of both visual and sequential channels. Our key observations are summarised as follows.
\begin{enumerate}
    \item From \figref{fig:rank observation} (a) and (b), higher rank representations tend to achieve higher F1 score. But due to the softmax capacity, significant performance drops can be seen in visual (after range 8-11) and sequential channels (after range 4-7), which confirms the existence of bottleneck. In other words, the softmax function cannot retain high-rank information for long complex videos.
    \item From the distribution of video numbers in \figref{fig:rank observation} (c) and (d), many video representations fall out of high-rank range (0-7) after softmax. According to the last observation, these videos are prone to getting lower performances.
    \item The softmax bottleneck problem is more severe on sequential attention, which indicates the changes between frames are the key missing information that results in the lower rank.
\end{enumerate}

Motivated by the above insights and inspired by the work of Yang \etal \cite{yang2017mos}, we come up with a \textbf{Mixture of Attention layer} (MoA) to alleviate the softmax bottleneck issue. We propose the Associated Query $\hat{\boldsymbol{Q}} = tanh(\boldsymbol{W}^{\hat{Q}}\boldsymbol{Q})$, where $\boldsymbol{W}^{\hat{Q}}$ is the Associated Query parameter. The idea is to capture the second-order changes between queries so that both complex and simple contents can be represented in a more smoothed attention representation.
The conditional attention distribution is defined as:

\begin{equation}
\begin{split}
       P(x \vert c) = &\sum_{t=1}^{T}\frac{\text{exp}(\mathcal{F}_{Scale}(\boldsymbol{K}_{c,t},\boldsymbol{Q}_{c,t}))}{\sum_{t=1}^{T}\text{exp}\mathcal{F}_{Scale}(\boldsymbol{K}_{c,t},\boldsymbol{Q}_{c,t})}\boldsymbol{\hat{A}}_{c,t}~\com \\ &s.t.\sum_{t=1}^{T}\boldsymbol{\hat{A}}_{c,t}=1~\com   
\end{split}
\label{eq:conditionalad}
\end{equation}
\begin{equation}
      where~\boldsymbol{\hat{A}} = \mathcal{F}_{Softmax}(\boldsymbol{K},\hat{\boldsymbol{Q}})~\com
      \label{eq:associatedad}
\end{equation}
In \eqnref{eq:associatedad}, $\boldsymbol{\hat{A}} \in T \times T $ is the associated attention distribution.
Thus, MoA formulates the conditional attention distribution as:
\begin{equation}
	\boldsymbol{A}_{moa} = \boldsymbol{A} \boldsymbol{\hat{A}}^T~\com
	\label{eq:moa}
\end{equation}
where $\boldsymbol{A}_{moa}  \in \mathbb{R}^{T \times T}$. In \eqnref{eq:dot-product}, due to the large degree of high dimensional $\boldsymbol{K}^T\boldsymbol{Q}$, scaling factor $\frac{1}{\sqrt{D_a}}$ is used to prevent the potential small gradient suffered by softmax. 
As $\boldsymbol{A}_{moa}$ is a non-linear function of the attention distribution, $\boldsymbol{A}_{moa}$ can be arbitrarily higher rank than standard self-attention structure $\boldsymbol{A}$. Thus the output of the mixture of attention network $\boldsymbol{Z} = \boldsymbol{V}\boldsymbol{A}_{moa}$ now can break the bottleneck problem. In \figref{fig:rank observation}, after applying the MoA, we can see a large proportion of videos fall into the 0-3 high rank range compared that of traditional softmax. Also, videos especially with lower ranks ($\mathcal{D}>11$) can be predicted with higher F1 scores. The performance of the sequential channel is boosted, which indicates that all of the previous softmax representations missed high rank information. The smoothed performance drop and increased number of high rank videos serve as strong evidence to manifest the Softmax Bottleneck has been resolved by proposed MoA.

 \begin{figure}[t]
 \centering
 \includegraphics[width=0.9\columnwidth]{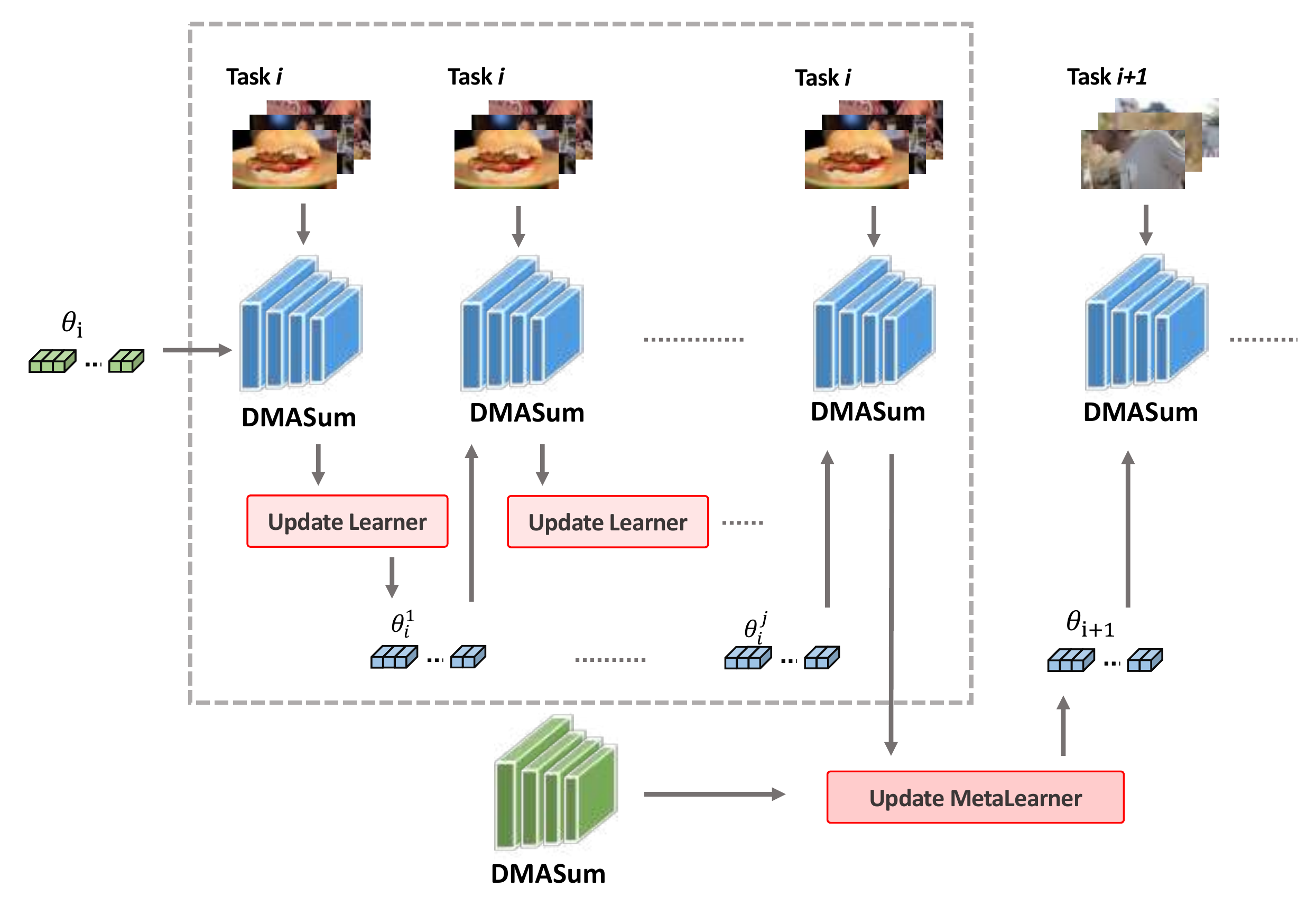}  
 \caption{Overview of the $i^{th}$ iteration for update $\theta_{i}$ to $\theta_{i+1}$. There are two stages in this update process. The middle part shows the stage about how the Learner updates $\theta_{i}$ to $\theta_{i}^{m}$ by iterating $m$ times. The outside parts shows the stage about how the Meta Learner updates $\theta_{i}$ to $\theta_{i+1}$.}
 \label{fig:meta}
 \end{figure}

Besides, the DMASum utilizes stacked mixture of attention networks, and in each layer we employ residual dropout connection \cite{he2016resnet} for allowing gradients to flow through a network directly and layer normalizaion \cite{ba2016layernorm} for normalizing the inputs across the features. Overall, the $n^{th}$ layer output can be defined as:

\begin{equation}
	\boldsymbol{Z}_n = \mathcal{F}_{Normalize}(\mathcal{F}_{Attention}(\boldsymbol{Z}_{n-1}) \oplus \boldsymbol{Z}_{n-1})\com
	\label{eq:attention}
\end{equation}
where $\mathcal{F}_{Normalize}$ denotes as layer normalization, $\mathcal{F}_{attention}$ represents the attention layer and $\oplus$ represents the residual connection.




\subsection{Single-Video Meta Learning}
The key motivation to introduce Meta Learning is to improve the model generalization when the dataset of video summarization is small. Different from gradient descent, the $MetaLearner$ is updated by weighted parameters of $Learner$ in subtasks, which can be formularized as:
\begin{equation}
	Learner^{*} = MetaLearner(Learner(\tau_{i}))
\end{equation}
where $\tau_{i}$ denotes $i^{th}$ video, $Learner$ and $Meta Learner$ means the DMASum model in meta learning. We first employ the MAML \cite{finn2017maml} due to its flexibility and superior performance but did not achieve expected results. Our observation is that in the video summarization context each video has its own latent mechanism that is not shared by different videos. Therefore, we propose a \textit{Single-Video Meta Learning} rule to refrain the learner by only one video at each task. The process is shown as \figref{fig:meta}.

There are two stages of each epoch in this meta learning strategy.
Firstly, to train the task $\tau_i$, the $Learner$ updates the parameter $\theta_{i}$ by traditional gradient descent. And, the $Learner$ trains the task in a set number $m$ recurrently to explore its latent summarizing context. The equation of updating parameter $\theta$ is:
\begin{equation}
	\theta_{i}^{j} = \theta_{i}^{j-1} - \alpha\nabla\mathcal{L}_{i}^{j}(\mathcal{F}_{\theta_{i}^{j-1}}), \quad where j = 1 \dots m
\end{equation}
where $\alpha$ denotes learning rate and $\nabla$ denoted as the gradient, and $\mathcal{F}_{\theta}$ is the loss function on $i^{th}$ task.
After $j^{th}$ iteration, the $Meta Learner$ updates the parameter $\theta_{i+1}$ by using the parameter $\theta_{i}^{m}$ of the $Learner$ by:
\begin{equation}
	\theta_{i} = \theta_{i-1} - \beta\nabla\mathcal{L}_{i}(\mathcal{F}_{\theta_{i}^{m}}),
	\label{eq:last step}
\end{equation}
where $\beta$ is the learning rate of the $Learner$. $\theta_{i}$ updated state of $Learner$ after the $j^{th}$ iteration in $Meta Learner$.
Overall, our meta learning is summarized in Algorithm \ref{ag:meta learning}.
Note that in the last step of the algorithm, we treat $\theta_{i}^{m}$ - $\theta_{i}$ as a gradient and plug it into Adam instead of simply updating $\theta_{i}$ in the direction $\theta_{i}^{m}$ - $\theta_{i}$.

\begin{algorithm}
\DontPrintSemicolon
 \SetKwInOut{Initialize}{Initialize}
 \tcc{$\theta$ : Parameter of $Learner$;}
 \tcc{$\alpha$ : Learning rate in $Learner$;}
 \tcc{$\beta$ : Learning rate in $Meta Learner$;}
 \tcc{$n$ : The number of videos;}
 \tcc{$m$ : Recurrent training $Learner$ number;}
 \tcc{$\mathcal{F}$ : the DMASum model;}
 \Initialize{$\theta$}
 \For{k = 1 to epoch number}{
   \For{i = 1 to n}
   {
       Sample video i as task $\tau_i$ \;
       \For{j = 1 to m}    
            { 
            	$\theta_{i}^{j} = \theta_{i}^{j-1} - \alpha\nabla\mathcal{L}_{i}^{j}(\mathcal{F}_{\theta_{i}^{j-1}})$
            }
        Update $\theta_{i+1} \leftarrow$ $\theta_{i}$ + $\beta$($\theta_{i}^{m}$ - $\theta_{i}$)
    }
 }
 \caption{Meta learning in DMASum}
 \label{ag:meta learning}
\end{algorithm}

\section{Experiments}
\subsection{Experiment Setup}

\noindent\textbf{Datasets.} 
We evaluate our model on two datasets: SumMe \cite{gygli2014summe} and TVSum \cite{song2015tvsum}. 
SumMe consists of 25 videos covering a variety of events, such as sports and cooking. The duration of each video varies from 1 to 6.5 minutes.
TVSum contains 50 videos downloaded from Youtube, which are selected from 10 categories. The video length varies from 1 to 10 minutes. 
Both datasets include ego-centric and third-person camera views, and the annotations were labeled by 25 human annotators. 
We also exploit two auxiliary datasets to augment the training data, where Open Video Project\footnote[1]{Open video project: https://open-video.org} (OVP) contains 50 videos and Youtube \cite{de2011youtube} contains 39 videos.

\noindent\textbf{Evaluation Metrics.} 
We follow the commonly used protocol from \cite{zhang2016vslstm} and converted the importance scores to shot-based summaries for both datasets, and the user annotations are changed from frame-level scores to key-shots scores using the kernel temporal segmentation (KTS) \cite{potapov2014kts} method, which can temporally segment a video into disjoint intervals. 
We then compute the harmonic mean F-score as the evaluation metric. In addition, according to the recent evaluation protocol \cite{otani2019rethinking}, we apply Kendall's $\tau$ \cite{kendall1945ken} and Spearman's $\rho$ \cite{zwillinger1999spe} correlation coefficients for comparing the ordinal association between generated summaries and the ground truth (i.e. the relationship between rankings). 
Also, they provided correlation curves to visualize the predicted importance score ranking with respect to the reference annotations, i.e., when the predicted importance scores are perfectly concordant with averaged human-annotated scores, the curve lies on the upper bound of the light-blue area. Otherwise, the curve coincides with the lower bound of the area when the ranking of the scores is in reverse order of the reference.

\noindent\textbf{Evaluation Settings.} 
Following \cite{zhang2016vslstm},  we conducted the experiments under three settings. 
(1) Canonical (C): we used the standard 5-fold cross-validation (5FCV) for SumMe and TVSum datasets. 
(2) Augmented (A): we used OVP and YouTube datasets to augment the training data in each fold under the 5FCV setting. 
(3) Transfer (T): we set a target testing dataset, e.g., SumMe or TVSum, and used the other three as the training data.

\noindent\textbf{Implementation details.} 
To be consistent with existing methods, the 1024 dimensional visual features extracted from the $pool5$ layer of the GoogLeNet \cite{szegedy2015googlenet} are used for training. 
To extract the temporal features, we design a Bi-LSTM model in the proposed network, as a two-layer LSTM with 512 hidden units per layer. 
For each attention layer, we set the attention dimension as 1024. 
We stack four attention layers for visual feature attention pipeline, and two layers for the sequential feature attention pipeline. 
The score layer consists of two fully-connected layers with 1024 hidden units.
For Single-video Meta Learning, we set the learning rate of $Learner$ as $\num{3e-5}$ and the learning rate of $MetaLearner$ as $\num{6e-5}$.
Moreover, the recurrent training Learner number is set as 3 and 5 in SumMe and TVSum datasets respectively.
During the test, we follow the strategy of prior work \cite{zhang2016vslstm,mahasseni2017sumgan,zhou2018dlsum} to generate the summary. 
In addition, we employ the ADAM optimizer to train our network and the hyperparameters are optimized via cross-validation.

\subsection{Quantitative Evaluation}
We first compare our method with state-of-the-art supervised approaches in three evaluation settings. Then, we re-implement the VS-LSTM, SUM-GAN, and DR-DSN models, and quote results for other methods from \cite{yuan2019cyclesum,jung2019csnet,ji2019autoae,wei2018sasum,he2019unsupervised,li2019MetaLTDVS,otani2019rethinking}. An in-depth ablation study is then provided to better understand of our DMASum. 

\begin{table}
\centering
\caption{F1-score (\%) of DMASum with state-of-the-art approaches on both SumMe and TVSum dataset.}
\begin{tabular}{ | l | c | c | }
  \hline			
  Method & SumMe & TVSum \\\hline\hline
  DPP-LSTM \cite{zhang2016vslstm} & 38.6 & 54.7 \\
  SASUM \cite{wei2018sasum}& 45.3 & 58.2 \\
  SUM-GAN \cite{mahasseni2017sumgan}& 41.7 & 54.3 \\
  Cycle-SUM \cite{yuan2019cyclesum}& 41.9 & 57.6 \\
  DR-DSN \cite{zhou2018dlsum}& 42.1 & 58.1 \\
  MetaL-TDVS \cite{li2019MetaLTDVS} & 44.1 & 58.2 \\
  ACGAN \cite{he2019unsupervised}& 46.0 & 58.5 \\
  CSNet \cite{jung2019csnet}& 51.3 & 58.8 \\
  M-AVS \cite{ji2019autoae}& 44.4 & 61.0 \\\hline\hline
  \textbf{DMASum} & \textbf{54.3} & \textbf{61.4} \\
  \hline
\end{tabular}
\label{tab:results all}
\end{table}

\begin{table}
\centering
\caption{Rank-order correlation coefficients computed between predicted importance scores by different models and human-annotated scores on both SumMe and TVSum datasets using Kendall's $\tau$ and Spearman's $\rho$ correlation coefficients.}
\begin{tabular}{ | l | c | c | c | c | }
  \hline			
  \multirow{2}{*}{Method} & \multicolumn{2}{c|}{SumMe} & \multicolumn{2}{c|}{TVSum}\\
  \cline{2-5}
  & $\tau$ &  $\rho$ &  $\tau$ & $\rho$  \\
  \hline\hline
  Random & 0.000 & 0.000 & 0.000 & 0.000 \\
  DPP-LSTM \cite{zhang2016vslstm}& - & - & 0.042 & 0.055 \\
  SUM-GAN \cite{mahasseni2017sumgan}& 0.049 & 0.066 & 0.024 & 0.031 \\
  DR-DSN \cite{zhou2018dlsum}& 0.028 & -0.027 & 0.020 & 0.026 \\
  Human & \textbf{0.227} & \textbf{0.239} & 0.178 & 0.205 \\\hline\hline
  \textbf{DMASum} & 0.063 & 0.089 & \textbf{0.203} & \textbf{0.267}   \\
  \hline
\end{tabular}
\label{tab:correlation coeeficients}
\end{table}

\noindent\textbf{Comparison with State-of-the-art Methods.}
Our main comparison with state-of-the-art methods is summarized in \tabref{tab:results all}. The compared methods can be mainly categorized into LSTM, GAN, Attention, and meta learning models. M-AVS \cite{ji2019autoae} and ACGAN \cite{he2019unsupervised} are based-on attention models and MetaL-TDVS \cite{li2019MetaLTDVS} is based on meta learning. It can be seen that DMASum outperforms other approaches on both datasets consistently. The F1-score results can reflect that our attention mechanism with meta learning can better predict importance scores. 

\begin{figure}[t]
\centering
    \begin{subfigure}[b]{0.25\textwidth}
        \centering
        \includegraphics[width=0.95\textwidth]{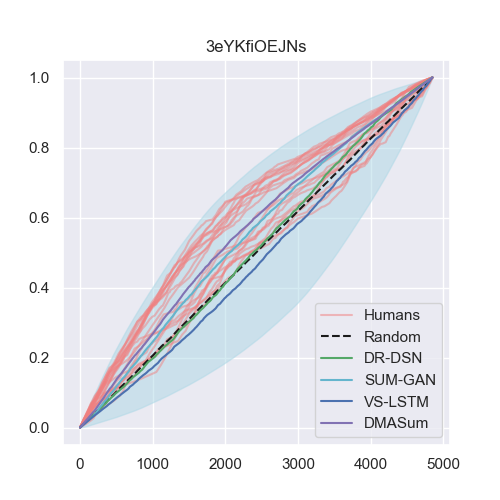}
    \end{subfigure}%
    \begin{subfigure}[b]{0.25\textwidth}
        \centering
        \includegraphics[width=0.95\textwidth]{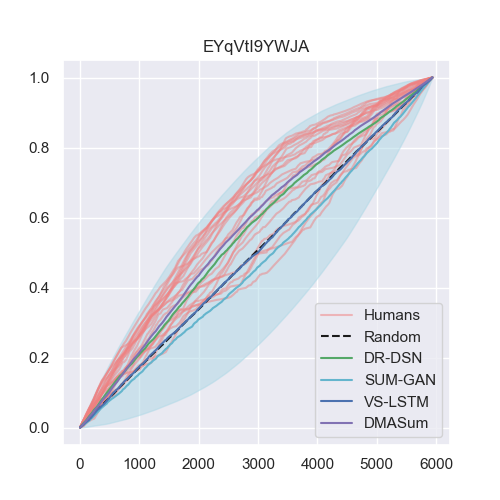}
    \end{subfigure}%
    \caption{Example correlation curves produced for two videos from the TVSum dataset (3eYKfiOEJNs and EYqVtl9YWJA are video ids). The red lines represent correlation curves for 25 human annotators and the black dashed line is the expectation for a random importance score. The magenta curve shows the corresponding result.}
\label{fig:correlation curves}
\end{figure}

We also evaluate our DMASum by using the most recent rank-order statistics \cite{otani2019rethinking}. The new evaluation matrix can also consider the frame dependencies and annotator consistency so as to reflect the true importance better. Because, F1 score can partially reflect the consistency between prediction and importance scores due to large variations in segment length (i.e. two-peak, KTS, and randomized KTS). The correlation coefficients (Kendall's $\tau$ and Spearman's $\rho$) can be used to measure the similarity between the implicit ranking provided by the frame-level importance score of the generated frame annotation and the human annotation. 
From \tabref{tab:correlation coeeficients}, We can see the correlation coefficients given by DMASum are significantly higher than other state-of-the-art models. More importantly, the performance on the TVSum dataset (0.233 and 0.267) is even better than human annotators (0.205 and 0.267). We believe it is because the dual-channel attention mechanism itself is simulating human behavior and memorizing visual and sequential sources of information and the meta learning method could learn the latent mechanism of summarizing a video story. Also, different human annotators might pay different attention to the given video. 
Our model can summarize the information from multiple human annotators so that the learned attention-based model is moderated and can achieve better consistency. 
\figref{fig:correlation curves} demonstrates two examples of the correlation coefficients. Curves above the random importance scores in the black dash are positive with better consistency. Our model achieves averaged performance among all human annotators and outperforms the other compared methods.

\begin{table}
\centering
\caption{F1-score (\%) of ablation study on SumMe and TVSum datasets. There are five ablation models: DMASum$_{wom}$ (without meta learning strategy), DMASum$_{softmax}$ (with standard softmax function in self-attention network), DMASum$_v$ (without sequential channel), DMASum$_s$ (without visual channel), DMASum$_b$ (with multiple videos in a batch), and DMASum$_{maml}$ (with MAML)}
\begin{tabular}{ | l | c | c | }
  \hline			
  Method & SumMe & TVSum \\\hline\hline
  DMASum$_{wom}$ & 51.6 & 60.6 \\
  DMASum$_{softmax}$ & 50.6 & 60.1 \\
  DMASum$_v$ & 53.2 & 60.5 \\
  DMASum$_s$ & 53.3 & 61.0 \\
  DMASum$_b$ & 51.3 & 60.0 \\
  DMASum$_{maml}$ & 49.3 & 59.2 \\
  \textbf{DMASum} & \textbf{54.3} & \textbf{61.4} \\
  \hline
\end{tabular}
\label{tab:ablation study}
\end{table}

\subsection{Ablation study.}
The success of our DMASum ascribes to both the framework design and technical improvement in each module. To analyze the effect of each component in DMASum, we conduct six ablation study models including DMASum without single video meta learning (DMASum$_{wom}$), DMASum with standard softmax function in self-attention network (DMASum$_{softmax}$), DMASum without sequential channel (DMASum$_v$), DMASum without visual channel (DMASum$_s$), DMASum$_b$ is developed with the batch version of Reptile, and DMASum is designed with MAML (DMASum$_{maml}$). Results are summarised in Table \ref{tab:ablation study}, from which we can understand the following questions.

\begin{figure}[t]
\centering
    \begin{subfigure}[b]{0.25\textwidth}
        \centering
        \includegraphics[width=0.95\textwidth]{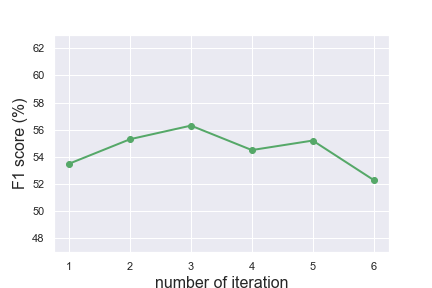}
        \caption{SumMe Dataset}
    \end{subfigure}%
    \begin{subfigure}[b]{0.25\textwidth}
        \centering
        \includegraphics[width=0.95\textwidth]{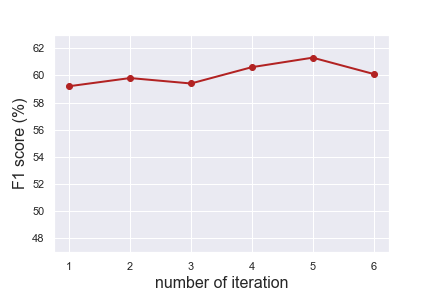}
        \caption{TVSum Dataset}
    \end{subfigure}%
\caption{Different recurrent training Learner number with respect to the F1-score (\%) in DMASum on both SumMe and TVSum datasets.}
\label{fig:meta learning number}
\end{figure}

\begin{figure*}[t]
    \centering
    \begin{subfigure}[b]{\textwidth}
        \centering
        \includegraphics[width=0.95\textwidth]{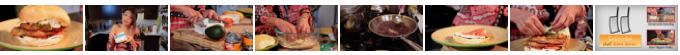}
        \caption{Example frames from video 37 in TVSum (indexed by \cite{song2015tvsum})}
    \end{subfigure}%
    \newline
    \begin{subfigure}[b]{0.5\textwidth}
        \centering
        \includegraphics[width=0.95\textwidth]{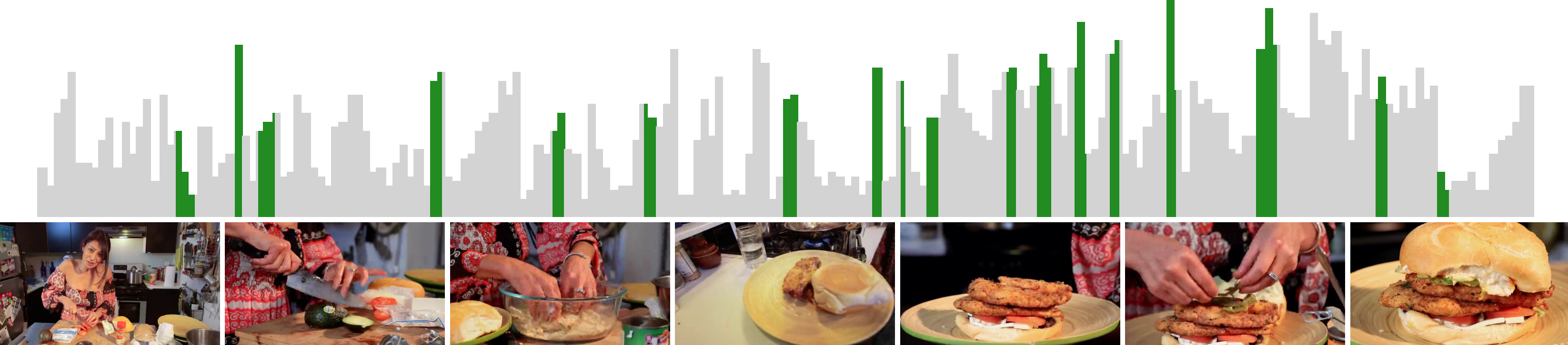}
        \caption{DMASum}
    \end{subfigure}%
    \begin{subfigure}[b]{0.5\textwidth}
        \centering
        \includegraphics[width=0.95\textwidth]{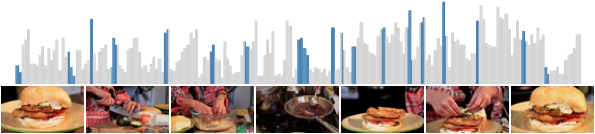}
        \caption{VS-LSTM}
    \end{subfigure}%
    \newline
    \begin{subfigure}[b]{0.5\textwidth}
        \centering
        \includegraphics[width=0.95\textwidth]{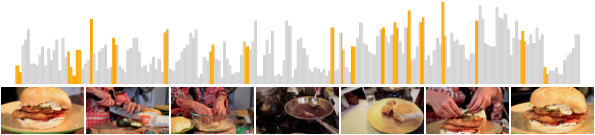}
        \caption{SUM-GAN}
    \end{subfigure}%
    \begin{subfigure}[b]{0.5\textwidth}
        \centering
        \includegraphics[width=0.95\textwidth]{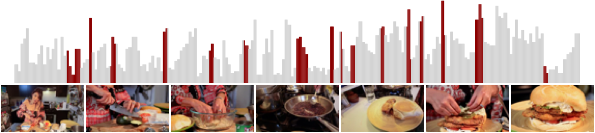}
        \caption{DR-DSN}
    \end{subfigure}%
    \medskip
    \vspace{-2ex}
    \caption{Quantitative results of different approaches for video 16 in TVSum. In (b) to (e), the light-gray bars represent the ground truth importance scores, and the colored bars correspond to the selected frames by different methods.}
    \label{fig:qualitative}
\end{figure*}

\noindent\textbf{The Basis of Self-attention Architecture} provided the initial performance boost. By removing our meta learning module, we can make a straight comparison with state-of-the-art DPP-LSTM \cite{zhang2016vslstm} and M-AVS \cite{ji2019autoae} which are using no attention and normal attention modules. Averaged improvement is around 5\% to 10\%. Note that M-AVS is slightly better than our method on the TVSum dataset due to their extra autoencoder architecture.

\noindent\textbf{The Softmax Bottleneck problem} results in severe performance gaps. By replace the MoA back to traditional softmax function, the performance drops 3.7\% and 1.3\% respectively on the two datasets. A more detailed analysis has been discussed in Section 3, from where we can see the problem is more critical when video contents are long and complex, involving rich sequential information.

\noindent\textbf{Visual vs Sequential Representation.} By comparing the performance of DMASum$_v$ or DMASum$_s$, we can observe that: 1) In TVSum dataset, the DMASum$_v$ gained a slightly better performance than DMASum$_s$. 2) The performance in SumMe dataset benefits more from the sequential channel. The self-attention network can effectively connect visual features from frames and the sequential information for the whole story line and thus our combined DMASum achieves better results.

\noindent\textbf{The Necessity of Meta Learning.} Removing the meta learning can heavily affect the performance by 2.7\% on SumMe dataset. The key reason is that SumMe is a relatively small dataset. This observation serves as strong evidence to validate the motivation and necessity of our meta learning module.
 
\noindent\textbf{MAML, Batch, and Single Video Meta Learning.} The Single Video rule is the key finding that distinguish it from meta learning in other applications, e.g. few-shot learning. This is due to a video itself is rich and complex. By increasing each meta learning task from one video to three in a batch, the performance of DMASum$_b$ drops 3\% and 1.4\% with the clearly slowed training process. In addition, we can see that the performance of our proposed meta learning strategy is better than the batch version of the Reptile strategy, and the batch version of the Reptile strategy is time-consuming during the training process. The efficiency of Single-Video rule is also validated by comparing it to DMASum$_{maml}$.

\noindent\textbf{Number of Recurrent Learning.}
In a controlled experiment, we observe that when the recurrent training Learner number is 3 for SumMe Dataset and 5 for the TVSum dataset, the F-score reaches the highest shown from \figref{fig:meta learning number}.
Which means, the Learner might not learn the summarizing mechanism when the number is too low, and when the number is too high, the Learner might overfit the current video. In this paper, the number of recurrent training is automatically chosen by using the standard 5-fold cross validation.

\noindent\textbf{Comparison under Different Settings.}
Another approach to examining the model generalization is to investigate its performance under different task settings. \tabref{tab:results settings} shows the experimental results of the comparison between the DMASum and cited results of state-of-the-art approaches in canonical, augmented and transfer settings.
Note that even though the performance of our model in augmented and transfer settings are partially better than the best results. We observe that the given importance scores in Youtube and OVP datasets are either 0 or 1. However, the DMASum is learning by the importance scores within the range of zero to one from SumMe and TVSum datasets. Such discrepancy of importance score format in both Youtube and OVP datasets would cause the meta learning strategy to be ineffective or even counterproductive because our model is not tailored to handle the discrepancy in labels. Thus in the future, we can improve our framework to adapt to this situation. But on the positive side, our DMASum is still capable in both augmented and transfer settings and achieves comparable results to that of state-of-the-art models despite the above difficulties.
\begin{table}
\centering
\caption{F-score (\%) of approaches in canonical, augmented and transfer settings on SumMe and TVSum datasets.}
\vspace{-2ex}
\begin{tabular}{ | l | c | c | c | c | c | c | }
  \hline			
  \multirow{2}{*}{Method} & \multicolumn{3}{c|}{SumMe} & \multicolumn{3}{c|}{TVSum}\\
  \cline{2-7}
  & C & A & T & C & A & T \\
  \hline\hline
  DPP-LSTM \cite{zhang2016vslstm}& 38.6 & 42.9 & 40.7 & 54.7 & 59.6 & 58.7 \\
  SUM-GAN \cite{mahasseni2017sumgan}& 41.7 & 43.6 & - & 54.3 & \textbf{61.2} & - \\
  DR-DSN \cite{zhou2018dlsum}& 42.1 & 43.9 & 42.6 & 58.1 & 59.8 & 58.9 \\
  CSNet \cite{jung2019csnet}& 51.3 & 52.1 & 45.1 & 58.8 & 59.0 & 59.2 \\\hline\hline
  \textbf{DMASum} & \textbf{54.3} & \textbf{54.1} & \textbf{52.2} & \textbf{61.4} & \textbf{61.2} & \textbf{60.5} \\
  \hline
\end{tabular}
\label{tab:results settings}
\end{table}

\subsection{Qualitative Evaluation}
To better illustrate the important frames selection of different approaches, we provide qualitative results for an exemplary video in \figref{fig:qualitative}, which tells a story of how to cook a burger. Overall, we can observe that all summaries generated by the different models can cover the intervals with high importance scores. Moreover, according to the figure, the summaries produced by both our DMASum and SUM-GAN contain more peaks, which proves that our proposed model can effectively capture key-frames from the original video. Also, the summary of our model is more sparse and much closer to the entire storyline, i.e., the different cooking stages, which means our meta learning strategy can learn the latent mechanism of summarizing a video.

\section{Conclusion}
We have presented the first work to introduce self-attention meta learning architecture to estimate the visual and sequential attentions jointly for video summarization. The self-attention formula was derived into a matrix factorization problem and key technical Softmax Bottleneck has been identified with both theoretical and empirical evidences. Our work also confirmed the importance of high-rank representation for video summarization tasks. A novel MoA module was proposed to replace the softmax, which can compare twice by query-key and self-query attentions. The Single-Video Meta Learning rule was designed and particularly tailored for video summarization tasks and significantly improved off-the-shelf Meta Learning, e.g. MAML. On two public datasets, our DMASum outperforms other methods in terms of both F1-score and achieved human-level performance using rank-order correlation coefficients. Future work could focus on further improve the generalisation for cross-dataset settings using an integrated framework.

\begin{acks}
Bingzhang Hu and Yu Guan are supported by Engineering and Physical Sciences Research Council (EPSRC) Project CRITiCaL: Combatting cRiminals In The CLoud (EP/M020576/1).
Yang Long is supported by Medical Research Council (MRC) Fellowship (MR/S003916/1).
\end{acks}

\balance
\bibliographystyle{ACM-Reference-Format}
\bibliography{mm}









\end{document}